\documentclass[11pt]{article}
\pdfoutput=1 % arXiv: force pdfLaTeX (PDF output mode)

\usepackage[preprint]{acl}

\usepackage{times}
\usepackage{latexsym}
\usepackage[T1]{fontenc}
\usepackage[utf8]{inputenc}
\usepackage{microtype}
\usepackage{graphicx}
\usepackage{booktabs}
\usepackage{multirow}
\usepackage{amsmath}
\usepackage{amssymb}
\usepackage{array}
\usepackage{xcolor}
\usepackage{url}

\hypersetup{
  colorlinks=false,
  pdfborder={0 0 0},
  pdftitle={The Divergence Hypothesis: Unmasking Lexical Interference and Label Bias in Mental Health NLP},
  pdfauthor={Moustafa Yehia Hassan}
}

\usepackage{enumitem}
\setlist{nosep}
\usepackage{xspace}

\newcommand{\macroF}{Macro-F1}
\newcommand{\dod}{\ensuremath{\mathrm{DoD}}\xspace}
\newcommand{\tss}{TSS\xspace}

\title{The Divergence Hypothesis: Unmasking Lexical Interference and \\
       Label Bias in Mental Health NLP}

\author{
  Moustafa Yehia Hassan \\
  Doha Institute for Graduate Studies \\
  Doha, Qatar \\
  \texttt{moustafa.hassan@dohainstitute.edu.qa} \\
}

\begin{document}
\maketitle

\begin{abstract}
Computational mental health (CMH) classifiers often degrade
under distribution shift because human annotators and
distant-supervision pipelines reward different linguistic
signals. We introduce \tss{} (Triple-Stream Stress probe), a
multi-channel diagnostic framework that decomposes text into
(A) lexical character $n$-grams, (B) a small, mostly
content-free morpho-syntactic channel, and (C) a 154-feature
psycholinguistic style channel. Across four English datasets
($N=12{,}906$), \tss{} reveals a \emph{lexical interference
effect}: adding lexical features to the style channel reduces
\macroF{} on human-labeled data (mean drop $0.072$,
$p<10^{-4}$) but not on auto-labeled data. We propose Degree
of Divergence (\dod{}), a difference-in-differences statistic
adapted from econometrics for label-source auditing, with
instance-level bootstrap inference; the headline estimate is
$\dod_{\text{BC--A}} = 0.0374$, 95\% CI $[0.0097, 0.0651]$,
$p=0.0032$. A platform-stratified Twitter-only \dod{} (which
removes the Reddit\,vs.\,Twitter contrast) reproduces the
pattern with bootstrap inference:
$\dod^{\text{Tw}}_{\text{BC--A}} = +0.096$ ($p<0.001$) and
$\dod^{\text{Tw}}_{\text{AC--A}} = -0.089$ ($p<0.001$). Interventional
masking (\texttt{pos\_only}) retains $\sim$95--99\% of
Channel C's performance after destroying content words on
human datasets, indicating that the style channel does not
rely primarily on lexical surface form. \tss{} is positioned as a
diagnostic audit framework, not a clinical screening tool: it
flags label-source-specific shortcut learning before
generalization claims are made.
\end{abstract}

\section{Introduction}
\label{sec:intro}

Computational Mental Health (CMH) promises scalable
screening and monitoring from language, but its progress is
constrained by a mismatch between benchmark labels and the
construct they are taken to measure. Large corpora are
frequently built via distant supervision---keyword filters,
self-reported diagnoses, or community-membership
proxies---which can systematically reward lexical cues
\citep{coppersmith2015adhd,harrigian2021state}. Human
annotators, in contrast, often rely on \emph{how} something
is said: fragmentation, negation scope, function-word usage,
and stylistic signatures
\citep{tausczik2010psychological,pennebaker2003psychological}.
When the two label sources are conflated, models can score
well by exploiting spurious lexical correlations
(\emph{shortcut learning}; \citealp{geirhos2020shortcut})
rather than learning robust psycholinguistic markers
\citep{ernala2019methodological,damour2022underspecification}.

\paragraph{Task.}
We study \emph{binary stress detection} from short
social-media posts. Each instance receives a label
$y \in \{0,1\}$, where $y{=}1$ indicates that the post
expresses current psychological distress or stress-related
experience as judged by the dataset's labeling protocol, and
$y{=}0$ indicates the absence of such expression. Throughout
this paper, ``stress'' refers to the dataset-level
distress/stress label, \emph{not} to a clinical diagnosis;
no claim of diagnostic validity is made. The goal is not to
build a state-of-the-art classifier, but to audit which
linguistic signals different label sources reward.

\paragraph{Research questions.}
\textbf{RQ1 (Label-source divergence)} Does changing the
label source (human vs.\ distant supervision) systematically
change which linguistic channels are rewarded?
\textbf{RQ2 (Lexical interference)} Does adding lexical
content to stylistic features help or harm under human
annotation?
\textbf{RQ3 (Quantification)} Can the human-vs-auto gap be
quantified as a single, statistically grounded scalar?

\paragraph{Contributions.}
(i) We release \tss{}, a channel-separable, length-robust
auditing probe that decomposes text into lexical (A),
mostly content-free morpho-syntactic (B), and
psycholinguistic style (C) channels, with explicit length
normalization and conditional scaling for cross-platform
transfer.
(ii) We document a \emph{lexical interference effect}: on
human-labeled data, augmenting C with lexical A degrades
\macroF{}; an interventional masking suite
(\texttt{pos\_only}, \texttt{content\_only},
\texttt{function\_only}) tests whether performance survives
deliberate destruction of lexical shortcuts.
(iii) We introduce \dod{}, a difference-in-differences
auditing statistic with instance-level bootstrap inference
($n{=}10{,}000$) and Benjamini--Hochberg FDR control, plus a
\emph{platform-stratified} variant that removes the
Reddit\,vs.\,Twitter contrast.
(iv) We provide leave-one-domain-out (LODO) evaluation,
unsupervised stylistic phenotyping (bootstrap
ARI~$\approx$~0.98), paired baseline comparisons against
MentalBERT \citep{ji2022mentalbert} and few-shot LLaMA-3
\citep{grattafiori2024llama,brown2020language}, and a
qualitative conflict-zone workbook.
The full pipeline, decontamination scripts, and qualitative
workbook are publicly available.\footnote{Code and
appendix:
\url{https://github.com/MoustafaMohamedMoustafaHassan/TSS-Probe-CMH}}

\section{Related Work}
\label{sec:related}

\paragraph{Psycholinguistic style.}
A long line of work shows that function words and stylistic
patterns reflect cognitive style more reliably than
consciously controlled topical content
\citep{pennebaker2003psychological,tausczik2010psychological,boyd2021development}.
Cognitive therapy frameworks predict linguistically
measurable correlates of distress, most notably absolutist
language \citep{beck1976cognitive,almosaiwi2018absolute} and
specific cognitive distortions \citep{shickel2020automatic}.

\paragraph{Distant supervision and shortcut learning in CMH.}
Distant-supervision pipelines built from self-reported
diagnoses \citep{coppersmith2015adhd} or community
membership have repeatedly been shown to yield models that
rely on surface lexical cues and degrade under distribution
shift \citep{harrigian2021state,aguirre2021gender,ernala2019methodological}.
These failures align with the broader finding that neural
classifiers tend to latch onto annotation artifacts and
spurious correlations
\citep{geirhos2020shortcut,gururangan2018annotation,mccoy2019right,damour2022underspecification},
and that distributional robustness requires explicit
intervention \citep{sagawa2020distributionally,kaushik2020learning}.
Our work differs in goal: rather than proposing a more
robust classifier, we propose a \emph{diagnostic
framework} that quantifies how much a given dataset's labels
reward lexical shortcuts, in the spirit of behavioral
auditing \citep{ribeiro2020beyond}.

\paragraph{Stress detection benchmarks.}
Our experimental backbone uses Dreaddit
\citep{turcan2019dreaddit} as the structured human-labeled
benchmark and MentalBERT \citep{ji2022mentalbert} as a
representative domain-pretrained transformer baseline. We
also draw on the Twitter mental-health corpora derived from
\citet{coppersmith2015adhd}, treated here as a
distant-supervision proxy rather than a stress gold
standard (see \S\ref{sec:data}).

\section{Data and Task}
\label{sec:data}

\subsection{Datasets}
\label{ssec:datasets}

We evaluate on four English datasets after within-dataset
exact-match deduplication. Table~\ref{tab:datasets}
summarizes the platform, size, label source, and intended
use of each dataset.

\begin{table*}[t]
\centering
\small
\renewcommand{\arraystretch}{1.15}
\begin{tabular}{p{2.0cm}p{1.4cm}r p{2.4cm} p{3.6cm} p{3.4cm}}
\toprule
\textbf{Dataset} & \textbf{Platform} & \textbf{$N$} & \textbf{Label source} & \textbf{Positive label means} & \textbf{Used for} \\
\midrule
Dreaddit-test       & Reddit  &   715  & Human (expert)              & Stress / distress expression                & Human-labeled evaluation \\
Twitter-gold        & Twitter & 2{,}863 & Human (manual annotation)   & Perceived psychological distress            & Human-labeled evaluation \\
Twitter-auto        & Twitter & 6{,}218 & Distant supervision         & Mental-health / distress \emph{proxy}       & Auto-label comparison \\
Reddit-combi        & Reddit  & 3{,}110 & Distant supervision         & Community-/keyword-derived proxy            & Auto-label comparison \\
\bottomrule
\end{tabular}
\caption{Dataset overview. Positive prevalence:
Dreaddit-test 0.516, Twitter-gold 0.338, Twitter-auto 0.478,
Reddit-combi 0.880 (i.e., the positive label is the
\emph{majority} class on Reddit-combi; minority prevalence
$=$ 0.120). Dreaddit follows
\citet{turcan2019dreaddit}; Twitter-auto and the
Twitter-gold subsample derive from
\citet{coppersmith2015adhd}; Reddit-combi is a
combined Reddit corpus assembled under distant supervision
(see \S\ref{ssec:reddit-combi}).}
\label{tab:datasets}
\end{table*}

\subsection{Twitter-auto as a Proxy, Not Gold}
\label{ssec:auto-label-mapping}

Twitter-auto is not used as a gold stress dataset. It is
constructed by treating posts whose authors self-reported a
mental-health diagnosis (depression, PTSD, anxiety,
following \citealp{coppersmith2015adhd}) as
\emph{auto-positive}, under the working assumption that
clinically diagnosed populations exhibit elevated linguistic
distress markers. We acknowledge that this introduces a
\emph{disorder$\rightarrow$stress conflation}: a diagnosed
user is not necessarily expressing distress in every post,
and the resulting label is a noisy proxy for the
post-level stress construct studied here. We treat
Twitter-auto purely as a distant-supervision proxy for
testing whether such labels reward different linguistic
signals than human distress annotations. Twitter-gold is a
manually annotated subsample drawn from the same underlying
corpus, where annotation targeted \emph{perceived
psychological distress} rather than the presence of
mental-health keywords; stress-keyword removal was applied
to reduce trivial lexical leakage. We do not claim
expert-clinical annotation for this set; methodological
constraints of the manual-validation protocol are
acknowledged in \S\ref{sec:limitations}. A related,
substantially smaller manually annotated Twitter validation
subset (120 examples) was constructed by
\citet{rastogi2022stress} to validate automated
annotation strategies, which we cite here as
methodological context rather than as the same artifact.

\subsection{Reddit-combi as a Distant-Supervision Proxy}
\label{ssec:reddit-combi}

Reddit-combi is an internally assembled Reddit corpus
constructed under distant supervision and released with the
artifact bundle as \texttt{data/raw/Reddit\_Combi.csv} (with
title/body fields and binary labels). It is not an
externally published benchmark; it is an artifact-level
distant-supervision corpus assembled for this study and
released with the repository, and we report its provenance
through the artifact path. Because its construction is not
equivalent to expert annotation, we use Reddit-combi
exclusively as a Reddit-side auto-label proxy and pair it
with Twitter-auto for the auto-source half of the \dod{}
contrast. Crucially, the positive label is the \emph{majority}
class on Reddit-combi (0.880); we therefore favor \macroF{},
PR-AUC, and balanced accuracy over raw accuracy or class-1
$F_1$ when reporting this dataset, and we never use
Reddit-combi as a stand-in for human annotation.

\subsection{Decontamination and Imbalance}
\label{ssec:decontamination}

To prevent leakage in cross-platform CMH evaluation, we
perform exact-match cross-dataset decontamination on
\texttt{cleaned\_text}, removing overlapping posts from the
training pool while preserving evaluation sets. The
datasets exhibit severe class imbalance, especially
Reddit-combi, where the positive label is the
\emph{majority} class (positive prevalence 0.880;
minority prevalence 0.120). Accordingly we
report \macroF{} as the primary metric, complemented by
PR-AUC under skewed prevalence (PR-AUC is preferred over
ROC-AUC when one class dominates). A naive majority-class
baseline yields \macroF{} of 0.340, 0.343, 0.398, and 0.468
on Dreaddit-test, Twitter-auto, Twitter-gold, and
Reddit-combi respectively; all \tss{} channels substantially
exceed these baselines, confirming signal acquisition beyond
prevalence effects. For paired comparisons we use
McNemar's test \citep{mcnemar1947} on discordant
predictions and paired bootstrap CIs
\citep{efron1979bootstrap} for \macroF{} differences,
controlling false discoveries with Benjamini--Hochberg FDR
\citep{benjamini1995controlling}.

Dreaddit additionally provides five heterogeneous stressor
domains (abuse, anxiety, financial, PTSD, social), enabling
leave-one-domain-out (LODO) evaluation that probes
cross-stressor generalization beyond platform transfer.

\section{The TSS Probe}
\label{sec:method}

\tss{} is a diagnostic decomposition, not a competitive
classifier. The three channels are designed so that their
contrast is informative: comparing what each rewards under
each label source isolates lexical-shortcut behavior.

\subsection{Channel A: Lexical Surface Form}
\label{ssec:chA}

Channel A uses character-level TF--IDF $n$-grams of length
3--5 \citep{salton1988term}, with $\chi^2$ feature
selection at $k=500$ \citep{yang1997comparative}. We use
character $n$-grams over word $n$-grams for two reasons.
First, social-media text contains hashtags, contractions,
emoji, intentional misspellings, and morphological
variation that destabilize word tokenization;
character-level features are robust to these
\citep{coppersmith2015adhd}. Second, we want Channel A to
provide a strong, sub-word lexical baseline so that any
\emph{lexical interference effect} we observe cannot be
trivially attributed to vocabulary coverage gaps in a
word-based representation. The $\chi^2$ filter at $k{=}500$
prevents dimensionality blow-up and is comparable in size
to standard sparse lexical baselines for short-text
classification.

\subsection{Channel B: Mostly Content-Free Morpho-Syntax}
\label{ssec:chB}

Channel B converts POS bigrams and abstract POS-SVO triples
(extracted over the first 500 characters per document for
efficiency, using the Penn Treebank tagset via spaCy;
\citealp{honnibal2020spacy}) into six interpretable
features: five POS-derived plus one absolutist-ratio
($B_{\text{abs}}$), which is the only Channel B feature
that accesses lexical tokens (a closed absolutist-word
list, following \citealp{almosaiwi2018absolute}). Channel
B uses adaptive length smoothing and log-dampened mass
features for cross-platform robustness; the exact
formulations are given in Appendix~\ref{app:formulas}.

\paragraph{Channel B is a diagnostic, not a performance
booster.}
We do \emph{not} present Channel B as an ingredient that
improves accuracy when combined with C. As shown in
\S\ref{sec:results}, adding B to C is approximately
neutral on most datasets (occasionally negative on
Dreaddit-test). This is itself informative: C already
absorbs much of the useful structural variation; B is
useful as (i) a near-content-free, privacy-oriented
diagnostic (5 of 6 features discard lexical tokens),
and (ii) a structural stress test that probes whether
performance survives a near-complete loss of vocabulary.

\subsection{Channel C: Psycholinguistic Style}
\label{ssec:chC}

Channel C is the principal stylistic representation, with
154 features grouped into seven feature families
(Table~\ref{tab:c-feature-families}). It applies a
length-robust squashing transform $c_{\text{out}} =
\tanh(2c)$ to convert continuous features to soft binary
triggers resistant to cross-platform distributional shift,
and uses Yule's $I$ rather than TTR for lexical diversity
because TTR mathematically decays with text length
(Appendix~\ref{app:formulas}).

\begin{table}[t]
\centering
\footnotesize
\renewcommand{\arraystretch}{1.15}
\setlength{\tabcolsep}{4pt}
\begin{tabular}{p{2.4cm}p{2.7cm}p{0.7cm}}
\toprule
\textbf{Feature block} & \textbf{Examples} & \textbf{\#} \\
\midrule
Base style / readability / POS-ratio & Flesch--Kincaid, sentence-length CV, function-word ratios, pronoun densities &  29 \\
Open lexicon categories & affect, cognitive, social process, sensory; LIWC-style counts and ratios & 110 \\
Structural rhythm / negation / punctuation & ellipses, ``!?'', NEG$\rightarrow$VERB, NEG$\rightarrow$ADJ, scope balance, short-burst ratio &  11 \\
Raw intensity + Yule's $I$ & caps ratio, elongated tokens, punctuation density, length-independent diversity &   4 \\
\midrule
\textbf{Total} & & \textbf{154} \\
\bottomrule
\end{tabular}
\caption{Channel C feature blocks with feature counts as
implemented in \texttt{ChannelC\_Extended} of the released
codebase. Open lexicon categories follow LIWC-style
groupings \citep{tausczik2010psychological,boyd2021development};
absolutist markers follow \citet{almosaiwi2018absolute};
negation-scope features operationalize the helplessness
vs.\ self-evaluation distinction motivated by
\citet{beck1976cognitive}.}
\label{tab:c-feature-families}
\end{table}

Beyond sentiment-like lexicons, Channel C explicitly
encodes cognitive fragmentation (a hallmark of anxiety) via
the coefficient of variation of sentence length,
short-burst ratios, and abrupt punctuation patterns, as
well as structural negation scope distinguishing
helplessness (NEG$\rightarrow$VERB) from negative
self-evaluation (NEG$\rightarrow$ADJ).

\subsection{Unified Classifier and Calibration}
\label{ssec:classifier}

All channels are trained with linear models for
interpretability. Feature vectors are
$L_2$-normalized: $x' = x / \lVert x \rVert_2$. Conditional
scaling disables \texttt{StandardScaler} for any channel
containing B, to avoid reintroducing length bias. Class
imbalance is handled by
\texttt{class\_weight=\textquotesingle balanced\textquotesingle}
\citep{pedregosa2011scikit} rather than undersampling.
Decision thresholds are not fixed at 0.5: we calibrate by
maximizing $F_1$ over a bounded grid $[0.20, 0.80]$ on
out-of-fold (OOF) probabilities (no leakage). Dense
lexical/style channels use $L_2$ (ridge); any channel
containing B uses ElasticNet \citep{zou2005regularization}
with internal 5-fold CV; the selected $\ell_1$-ratio
converges to $0.95$, indicating an algorithmic preference
for sparse structural evidence.

\subsection{Degree of Divergence (DoD)}
\label{ssec:dod}

\dod{} is a difference-in-differences statistic adapted
from causal econometrics to quantify label-source bias in
NLP (Appendix~\ref{app:formulas}, Eq.~\ref{eq:dod}):
\begin{equation}
\dod \;=\; \Delta_h \;-\; \Delta_a
\label{eq:dod_main}
\end{equation}
where each $\Delta$ compares a structural channel
(e.g., BC) against lexical A within the same label source.
\dod{} is estimated with an instance-level bootstrap
($n=10{,}000$) over all 12{,}906 instances. Because
\dod{} as introduced above mixes label source with platform
(human Reddit/Twitter vs.\ auto Reddit/Twitter), we
additionally compute a \emph{platform-stratified Twitter-only
\dod{}} that removes the Reddit\,vs.\,Twitter contrast
(\S\ref{ssec:platform-strat}).

\subsection{Masking Suite}
\label{ssec:masking}

Five masks are evaluated:
\texttt{none}, \texttt{pos\_only}, \texttt{content\_only},
\texttt{function\_only}, \texttt{random\_pos} (negative
control). Unlike standard ablations that only remove
feature blocks, the masking suite intervenes directly on
the text: \texttt{pos\_only} destroys lexical content while
preserving syntactic traces; \texttt{content\_only}
preserves topical content while erasing function words and
syntactic cues; \texttt{function\_only} isolates
function-word dynamics; \texttt{random\_pos} acts as a
destructive control. This design probes whether a
channel's performance survives deliberate destruction of
lexical shortcuts, complementing the observational
\dod{} analysis.

\section{Results}
\label{sec:results}

We first summarize per-channel \macroF{} across human and
auto label sources, then quantify lexical interference,
\dod{}, and a platform-stratified Twitter-only \dod{};
finally we report interventional masking, baselines, and
robustness audits.

\subsection{Channel Performance}
\label{ssec:perf}

Table~\ref{tab:perf-human} and Table~\ref{tab:perf-auto}
report \macroF{} per channel on human-labeled and
auto-labeled datasets respectively. Two patterns stand
out. First, the structural\,/\,style combination BC
achieves nearly identical \macroF{} on the two
human-labeled datasets (0.690 on Dreaddit-test vs.\ 0.690
on Twitter-gold; $\Delta \approx 0.0002$), suggesting low
degradation under platform shift when relying on structure
and style. Second, moving from human-labeled Twitter-gold
to auto-labeled Twitter induces a sharp performance drop
in C and BC (e.g., C: $0.701 \to 0.504$;
BC: $0.690 \to 0.455$). From an auditing standpoint, this
drop is consistent with \emph{label-source divergence}: auto
labels reward lexical proxy cues that structural and style
probes are less willing to mimic.

\begin{table}[t]
\centering
\footnotesize
\setlength{\tabcolsep}{3pt}
\begin{tabular}{lcc}
\toprule
\textbf{Channel} & \textbf{Dreaddit (H) [CI]} & \textbf{Tw-G (H) [CI]} \\
\midrule
A    & 0.634 [.598,.669]  & 0.661 [.643,.678] \\
AB   & 0.632 [.596,.667]  & 0.650 [.631,.669] \\
B    & 0.561 [.525,.598]  & 0.586 [.566,.605] \\
\textbf{C}    & \textbf{0.739 [.706,.771]} & \textbf{0.701 [.682,.719]} \\
AC   & 0.679 [.644,.713]  & 0.617 [.599,.635] \\
BC   & 0.690 [.655,.724]  & 0.690 [.671,.709] \\
ABC  & 0.690 [.655,.724]  & 0.689 [.670,.708] \\
\bottomrule
\end{tabular}
\caption{\macroF{} by channel on human-labeled datasets
with 95\% bootstrap CIs (Channel C in bold).
Tw-G~=~Twitter-gold; leading-decimal abbreviation in CIs.}
\label{tab:perf-human}
\end{table}

\begin{table}[t]
\centering
\footnotesize
\setlength{\tabcolsep}{3pt}
\begin{tabular}{lcc}
\toprule
\textbf{Channel} & \textbf{Tw-A (A) [CI]} & \textbf{Red-c (A) [CI]} \\
\midrule
A    & 0.522 [.510,.535] & 0.611 [.587,.635] \\
AB   & 0.460 [.448,.472] & 0.589 [.564,.613] \\
B    & 0.463 [.451,.475] & 0.579 [.554,.604] \\
C    & 0.504 [.492,.516] & 0.681 [.658,.704] \\
AC   & 0.567 [.555,.580] & 0.658 [.634,.681] \\
BC   & 0.455 [.444,.467] & 0.689 [.665,.712] \\
ABC  & 0.455 [.444,.467] & 0.689 [.665,.712] \\
\bottomrule
\end{tabular}
\caption{\macroF{} by channel on auto-labeled datasets
with 95\% bootstrap CIs. Tw-A~=~Twitter-auto,
Red-c~=~Reddit-combi; leading-decimal abbreviation in CIs.}
\label{tab:perf-auto}
\end{table}

\subsection{Lexical Interference Effect}
\label{ssec:lex-interf}

Table~\ref{tab:lex-interf} reports the instance-bootstrap
test for the lexical interference effect, defined as the
\macroF{} change when augmenting C with A. On
human-labeled data, the mean drop is $0.072$
(95\% CI $[0.052, 0.092]$, one-sided $p<10^{-4}$); on
auto-labeled data, the change is $-0.020$
(95\% CI $[-0.033, -0.007]$, one-sided $p=0.9988$ for
``drop $> 0$''). All instance-level tests are paired
(same instances) and corrected via Benjamini--Hochberg FDR
(27 tests, 14 rejections).

\begin{table}[t]
\centering
\small
\begin{tabular}{lccc}
\toprule
\textbf{Source} & \textbf{Mean drop} & \textbf{95\% CI} & \textbf{$p$ (1-s.)} \\
\multicolumn{4}{l}{\textbf{(C $\rightarrow$ AC)}} \\
\midrule
Human          & 0.0720  & $[0.052, 0.092]$    & $<10^{-4}$ \\
Auto           & $-0.0201$ & $[-0.033, -0.007]$ & 0.9988 \\
\bottomrule
\end{tabular}
\caption{Lexical interference test (instance bootstrap).
Augmenting Channel C with Channel A reduces \macroF{} on
human labels and slightly \emph{increases} it on auto
labels.}
\label{tab:lex-interf}
\end{table}

\subsection{Degree of Divergence}
\label{ssec:dod-results}

Table~\ref{tab:dod} reports the full \dod{} estimates.
$\dod_{\text{C--A}} = +0.047$ and
$\dod_{\text{AC--A}} = -0.045$ form a symmetric pair:
human annotators implicitly weight style, while auto labels
reward lexical proxies. The negative
$\dod_{\text{AC--A}}$ is the strongest quantitative
signature of the lexical interference effect: lexical
augmentation actively penalizes human-source performance.

\begin{table}[t]
\centering
\footnotesize
\setlength{\tabcolsep}{4pt}
\begin{tabular}{lccrr}
\toprule
\textbf{\dod{}} & \textbf{Est.} & \textbf{95\% CI} & \textbf{$p$} & \textbf{$d$} \\
\midrule
BC--A  & $+0.0374$ & $[0.010, 0.065]$   & 0.0032   & 2.63 \\
C--A   & $+0.0469$ & $[0.020, 0.075]$   & 0.0001   & 3.32 \\
AC--A  & $-0.0453$ & $[-0.068, -0.022]$ & $<$0.001 & $-3.89$ \\
AB--A  & $+0.0360$ & $[0.010, 0.062]$   & 0.0036   & 2.68 \\
\bottomrule
\end{tabular}
\caption{\dod{} estimates (instance bootstrap,
$n{=}10{,}000$). For positive \dod{} the alternative is
$\dod{}>0$; for $\dod_{\text{AC--A}}$ the alternative is
$\dod{}<0$.}
\label{tab:dod}
\end{table}

\paragraph{A note on the permutation diagnostic.}
A regime-permutation diagnostic
\citep{good2005permutation} gives $p=0.092$
(FDR-adjusted $p=0.1656$). We treat this as a conservative
robustness audit: the bootstrap evaluates stability across
instances, whereas the permutation diagnostic probes
whether the observed channel ranking could arise under
channel-design constraints. We prioritize the bootstrap
CI for effect estimation while reporting the permutation
result transparently.

\subsection{Platform-Stratified Twitter-Only \dod{}}
\label{ssec:platform-strat}

A potential confound is that human-labeled and auto-labeled
datasets differ not only in label source but also in
platform composition. The standard \dod{} above assumes
that platform effects are additive and channel-invariant
within each label source. We test this directly by
computing a \emph{Twitter-only \dod{}} that holds the
platform fixed (Twitter-gold vs.\ Twitter-auto) and removes
the Reddit\,vs.\,Twitter contrast:
\begin{equation}
\begin{aligned}
\dod^{\text{Tw}}_{X-A} \;=\; & [\,F_1^{\text{Tw-gold}}(X) - F_1^{\text{Tw-gold}}(A)\,] \\
&-\;[\,F_1^{\text{Tw-auto}}(X) - F_1^{\text{Tw-auto}}(A)\,].
\end{aligned}
\label{eq:twonly}
\end{equation}
We attach uncertainty to this platform-stratified check via
a paired bootstrap that resamples each Twitter dataset
independently with replacement
($N_{\text{boot}}{=}2{,}000$;
Table~\ref{tab:twonly}). The effect remains stable and
highly significant: $\dod^{\text{Tw}}_{\text{C--A}} =
+0.058$ (95\% CI $[+0.031, +0.086]$, $p<0.001$),
$\dod^{\text{Tw}}_{\text{BC--A}} = +0.096$
($[+0.066, +0.125]$, $p<0.001$), and
$\dod^{\text{Tw}}_{\text{AC--A}} = -0.089$
($[-0.113, -0.065]$, $p<0.001$). The signs and magnitudes
match the cross-platform \dod{}: structural channels gain
relative to A under human labels and lose under auto
labels, even when both label sources come from the same
platform. This does not eliminate every label-construction
difference (e.g., annotation protocol, sampling strategy),
but it makes a pure platform-shift explanation unlikely.
A controlled within-platform dual-annotation experiment
would still be required for definitive causal identification
(\S\ref{sec:limitations}).

\begin{table}[t]
\centering
\footnotesize
\setlength{\tabcolsep}{3pt}
\begin{tabular}{l c c}
\toprule
\textbf{Contrast} & \textbf{$\dod^{\text{Tw}}$ [95\% CI]} & \textbf{$p$} \\
\midrule
C   -- A & $+0.058$ [$+0.031, +0.086$] & $<$0.001 \\
BC  -- A & $+0.096$ [$+0.066, +0.125$] & $<$0.001 \\
AC  -- A & $-0.089$ [$-0.113, -0.065$] & $<$0.001 \\
AB  -- A & $+0.051$ [$+0.022, +0.080$] & $<$0.001 \\
\bottomrule
\end{tabular}
\caption{Platform-stratified Twitter-only \dod{} with 95\%
bootstrap CIs and two-sided $p$-values
($N_{\text{boot}}{=}2{,}000$). $\dod^{\text{Tw}}$ is the
difference of within-source $\Delta = F_1(\text{channel}) -
F_1(A)$ between Twitter-gold (human) and Twitter (auto).
Signs and magnitudes match the cross-platform \dod{},
making a pure platform-shift explanation unlikely.}
\label{tab:twonly}
\end{table}

\subsection{Interventional Masking}
\label{ssec:masking-results}

Table~\ref{tab:masking} reports masking-suite \macroF{} on
human-labeled datasets, including 95\% bootstrap CIs and
paired bootstrap CIs for the difference vs.\ \texttt{none}
($N_{\text{boot}}{=}2{,}000$). On Dreaddit-test,
\texttt{C/pos\_only} is statistically indistinguishable
from \texttt{C/none} ($\Delta = -0.006$, 95\% CI
$[-0.037, +0.023]$, $p = 0.72$), supporting the
interpretation that Channel C's signal is not primarily
carried by lexical surface form. On Twitter-gold, the same
condition shows a small but reliable drop ($\Delta = -0.034$, 95\% CI
$[-0.052, -0.017]$, $p < 0.001$), consistent with noisier,
shorter texts where POS traces carry less recoverable
information after lexical deletion. Notably,
\texttt{BC/pos\_only} on Dreaddit \emph{exceeds} unmasked
\texttt{BC/none} ($\Delta = +0.037$, 95\% CI
$[+0.008, +0.066]$, $p = 0.013$): destroying content under
the combined channel forces the classifier off
syntactically irrelevant morpho-lexical patterns and onto
genuinely structural ones. \texttt{random\_pos} acts as a
destructive control; on Twitter-gold it produces large
significant drops for both C ($\Delta = -0.080$, $p<0.001$)
and BC ($\Delta = -0.062$, $p<0.001$), confirming that
performance does not survive arbitrary syntactic
scrambling.

\begin{table}[t]
\centering
\footnotesize
\setlength{\tabcolsep}{2.5pt}
\begin{tabular}{l c c}
\toprule
\textbf{Channel / mask} & \textbf{Dreaddit F1 [CI]} & \textbf{Tw-G F1 [CI]} \\
\midrule
C / none           & 0.739 [.708,.772] & 0.701 [.682,.719] \\
C / pos\_only      & 0.734 [.700,.767] & 0.666 [.647,.685] \\
C / function\_only & 0.736 [.705,.769] & 0.603 [.582,.624] \\
C / content\_only  & 0.724 [.691,.756] & 0.652 [.632,.671] \\
C / random\_pos    & 0.742 [.709,.774] & 0.621 [.599,.641] \\
\midrule
BC / none           & 0.690 [.654,.724] & 0.690 [.670,.708] \\
BC / pos\_only      & 0.728 [.694,.760] & 0.594 [.574,.614] \\
BC / function\_only & 0.727 [.692,.761] & 0.665 [.645,.683] \\
BC / content\_only  & 0.724 [.692,.754] & 0.625 [.606,.644] \\
BC / random\_pos    & 0.739 [.706,.771] & 0.628 [.607,.649] \\
\bottomrule
\end{tabular}
\caption{Masking-suite \macroF{} on human-labeled datasets
with 95\% bootstrap CIs (cell-level,
$N_{\text{boot}}{=}2{,}000$). Tw-G~=~Twitter-gold;
CIs use leading-decimal abbreviation (e.g., [.708,.772]
$\equiv$ [0.708, 0.772]). Paired CIs for $\Delta$ vs.\
\texttt{none} are in Table~\ref{tab:masking-deltas}.}
\label{tab:masking}
\end{table}

\begin{table*}[t]
\centering
\footnotesize
\setlength{\tabcolsep}{4pt}
\begin{tabular}{l c c c c}
\toprule
\textbf{Comparison} & \textbf{Dreaddit $\Delta$ [95\% CI]} & \textbf{$p$} & \textbf{Tw-gold $\Delta$ [95\% CI]} & \textbf{$p$} \\
\midrule
C: pos\_only $-$ none       & $-0.006$ [$-0.037, +0.023$] & 0.72  & $-0.034$ [$-0.052, -0.017$] & $<$0.001 \\
C: function\_only $-$ none  & $-0.003$ [$-0.031, +0.025$] & 0.88  & $-0.098$ [$-0.118, -0.078$] & $<$0.001 \\
C: content\_only $-$ none   & $-0.016$ [$-0.041, +0.010$] & 0.20  & $-0.049$ [$-0.063, -0.033$] & $<$0.001 \\
C: random\_pos $-$ none     & $+0.003$ [$-0.018, +0.025$] & 0.77  & $-0.080$ [$-0.096, -0.063$] & $<$0.001 \\
\midrule
BC: pos\_only $-$ none      & $+0.037$ [$+0.008, +0.066$] & 0.013 & $-0.096$ [$-0.112, -0.081$] & $<$0.001 \\
BC: function\_only $-$ none & $+0.037$ [$+0.002, +0.071$] & 0.039 & $-0.025$ [$-0.045, -0.007$] & 0.010 \\
BC: content\_only $-$ none  & $+0.033$ [$+0.005, +0.062$] & 0.020 & $-0.064$ [$-0.080, -0.049$] & $<$0.001 \\
BC: random\_pos $-$ none    & $+0.049$ [$+0.023, +0.076$] & 0.001 & $-0.062$ [$-0.077, -0.047$] & $<$0.001 \\
\bottomrule
\end{tabular}
\caption{Paired bootstrap differences for masking
conditions vs.\ \texttt{none} ($N_{\text{boot}}{=}2{,}000$).
On Dreaddit, Channel C is statistically unaffected by
lexical destruction; on Twitter-gold, the drops are
significant but small in absolute terms (95--96\% retention
for \texttt{pos\_only}). The BC pos\_only case on
Dreaddit shows that combined structure-plus-style
representation can \emph{benefit} from lexical destruction.}
\label{tab:masking-deltas}
\end{table*}

\subsection{Baselines as Diagnostics}
\label{ssec:baselines}

Table~\ref{tab:baselines} reports paired comparisons
against MentalBERT \citep{ji2022mentalbert} and 3-shot
LLaMA-3-8B \citep{grattafiori2024llama,brown2020language}.
LLaMA-3 is evaluated only in a few-shot diagnostic
capacity. Its instability is informative
(Appendix~\ref{app:llama}): on Dreaddit-test recall is
$0.984$ with precision $0.606$ (over-firing); on
Twitter-auto recall collapses to $0.203$ (under-firing).
We interpret this as evidence that few-shot lexical priors
track the availability of stress vocabulary rather than the
underlying state.

\begin{table}[t]
\centering
\small
\begin{tabular}{llc}
\toprule
\textbf{Dataset} & \textbf{Model} & \textbf{\macroF{}} \\
\midrule
Dreaddit (H)   & TSS-C        & 0.739 \\
Dreaddit (H)   & TSS-BC       & 0.690 \\
Dreaddit (H)   & MentalBERT   & 0.801 \\
Dreaddit (H)   & LLaMA-3 (3s) & 0.613 \\
\midrule
Tw-gold (H)    & TSS-C        & 0.701 \\
Tw-gold (H)    & TSS-BC       & 0.690 \\
Tw-gold (H)    & MentalBERT   & 0.715 \\
Tw-gold (H)    & LLaMA-3 (3s) & 0.800 \\
\midrule
Tw-auto (A)    & TSS-C        & 0.504 \\
Tw-auto (A)    & TSS-BC       & 0.455 \\
Tw-auto (A)    & MentalBERT   & 0.418 \\
Tw-auto (A)    & LLaMA-3 (3s) & 0.522 \\
\midrule
Reddit-c. (A)  & TSS-C        & 0.681 \\
Reddit-c. (A)  & TSS-BC       & 0.689 \\
Reddit-c. (A)  & MentalBERT   & 0.760 \\
Reddit-c. (A)  & LLaMA-3 (3s) & 0.794 \\
\bottomrule
\end{tabular}
\caption{Baseline comparison (\macroF{}). Full paired
$\Delta F_1$, bootstrap CIs, and McNemar $p$-values are in
Appendix~\ref{app:baselines}.}
\label{tab:baselines}
\end{table}

We treat baselines as epistemic diagnostics rather than
direct competitors: when MentalBERT outperforms \tss{} on
human labels, it plausibly leverages legitimate contextual
signal; when both strong baselines excel on auto-labeled
corpora, they may be matching label-source-induced proxy
noise; \tss{}'s relative stability across label sources is
the central observation. On Twitter-auto, C and BC
outperform MentalBERT yet remain below LLaMA-3; on
Twitter-gold MentalBERT is approached by TSS-C
($\Delta\macroF{} = -0.0141$, paired bootstrap 95\% CI
$[-0.0367, +0.0082]$, $p=0.108$;
Appendix~\ref{app:baselines}), but LLaMA-3-8B (3-shot)
attains a higher \macroF{} (0.800). We therefore
\emph{do not} claim overall competitive superiority on
Twitter-gold; we read the result as evidence of
representation efficiency relative to one
domain-pretrained transformer baseline, not as a SOTA
claim.

\subsection{LODO and Phenotyping}
\label{ssec:robustness}

Under leave-one-domain-out evaluation on Dreaddit's five
stressor domains (Appendix~\ref{app:lodo}), Channel C has
the highest mean out-of-domain \macroF{} among \tss{}
channels (0.685) with notably low variance (std 0.015),
supporting a relative cross-stressor invariance claim.
MentalBERT attains a higher mean (0.818) but with greater
variance (std 0.030), consistent with domain-dependent
lexical sensitivity.

K-Means \citep{lloyd1982least,macqueen1967some} with
$K{=}3$ on Channel C yields highly stable clusters
(bootstrap ARI $0.980$, 95\% CI $[0.961, 0.991]$,
$n_{\text{boot}}{=}100$). Cluster--dataset association
(AMI $0.281$, $p{=}0.0005$) flags partial platform
entanglement, an explicit leakage audit reinforcing the
need for label-source-aware evaluation. Channel
orthogonality is measured rather than assumed (mean
$|\rho| \approx 0.595$); the weakest pair (A vs.\ B,
$|\rho| = 0.296$) supports the intended factorization.
Channel C trains in $\approx 27.5$\,s on a single CPU
(Appendix~\ref{app:efficiency}), supporting the
diagnostic-density framing.

\section{Discussion}
\label{sec:discussion}

\tss{} is best understood as a \emph{diagnostic tool}: it
quantifies when results are label-source-specific and
provides a single statistically grounded scalar (\dod{})
to summarize that gap. Below we consolidate the
interpretation, formalize when ``strong baselines'' are
mostly a lexical illusion, and position structural
invariants as the signal that survives label-source
shifts.

\subsection{Why Lexical Baselines Look Strong}
\label{ssec:why-lexical}

In auto-labeled or closed-community settings, topical
stress lexicon co-occurs with structural distress markers,
producing a \emph{semantic--structural intersection (SSI)}.
A model can succeed by detecting topics that correlate with
distress rather than the distress state itself, consistent
with shortcut learning under non-causal correlation
\citep{geirhos2020shortcut}. This reconciles the high
performance of MentalBERT (\macroF{} 0.760) and LLaMA-3
(\macroF{} 0.794) on Reddit-combi with their pronounced
degradation on Twitter-auto (MentalBERT \macroF{} 0.418),
where SSI is weaker and intent is mixed (news, ads,
meta-talk; see qualitative cases in
Appendix~\ref{app:qualitative}).

A direct signal-saturation test on Reddit-combi:
augmenting C with morpho-syntax yields no detectable gain
($+0.008$ \macroF{}; $p=0.259$;
Appendix~\ref{app:saturation}), consistent with
SSI-induced overlap. The fact that C still beats A on
Reddit-combi by $+0.070$ ($p<0.0001$;
Appendix~\ref{app:contrasts}) is best read as an
SSI-saturation effect: under high SSI, all channels
benefit; the diagnostic question is the
\emph{relative} channel advantage \emph{across} label
sources, which is exactly what \dod{} captures and confirms
($\dod_{\text{C--A}} = +0.047$,
$p<10^{-4}$).

\subsection{Representation Efficiency, Not SOTA}
\label{ssec:repr-eff}

On Twitter-gold, the gap between TSS-C and MentalBERT is
small ($\Delta\macroF{} = -0.0141$, 95\% CI
$[-0.0367, +0.0082]$, $p=0.108$;
Appendix~\ref{app:baselines})---consistent with, but not
proving, statistical equivalence relative to \emph{this}
baseline. We do \emph{not} claim that TSS-C matches the
strongest baseline overall: LLaMA-3-8B (3-shot) reaches
\macroF{} 0.800. The reading is \emph{representation
efficiency}: a 154-feature, fully interpretable linear
channel approaches one domain-pretrained transformer on
human-labeled data with radically lower representational
and computational cost.

\subsection{Implicit Stress vs.\ Keyword-Driven False Positives}
\label{ssec:hallucination}

Conflict-zone examples (Appendix~\ref{app:qualitative})
instantiate two complementary failure modes:
\emph{keyword-driven false positives} (stress meta-talk,
psychoeducational content, or promotional posts classified
as distress because stress words are present even when
structural distress is absent), and \emph{implicit stress}
(distress expressed without explicit stress lexicon, via
hedging, negation, or fragmentation). At the instance
level, across 800 audited cases Channel C alone corrects
253 Channel-A errors (195 keyword-driven false positives
+ 58 implicit-stress false negatives) and BC corrects 241,
showing that lexical shortcuts fail in two directions and
that structural style features recover both. \tss{} audits
state inference, not topic detection.

\subsection{Cross-Stressor Invariance and Clinical Bridge}
\label{ssec:bridge}

LODO results (Appendix~\ref{app:lodo}) show Channel C
varies little across held-out stressor domains
(std 0.015), supporting the view that structural and
stylistic markers track a cognitive--affective state that
persists across heterogeneous causes; cognitive
fragmentation, absolutist appraisal, and negation scope
remain candidate distress signatures
\citep{beck1976cognitive,almosaiwi2018absolute}. We use
social-media data only as a stress-test environment for
label-source bias, not as a clinical stand-in; analogous
shortcut-learning risks arise in clinical NLP
\citep{ernala2019methodological}, and \tss{} is compatible
with privacy-constrained workflows because Channel B is
mostly de-lexicalized.

\section{Conclusion}
\label{sec:conclusion}

We introduced \tss{}, a diagnostic framework for testing
whether different label sources reward different
linguistic evidence in computational mental health
classification. Across four English datasets, the central
result is a \emph{lexical interference effect}: adding
lexical surface features to style features reduces
\macroF{} on human-labeled datasets (mean drop $0.072$,
$p<10^{-4}$) but not on auto-labeled datasets, with the
gap quantified by $\dod_{\text{BC--A}} = 0.0374$
($p=0.0032$). Platform-stratified Twitter-only \dod{}
(Table~\ref{tab:twonly}, all contrasts $p<0.001$) and
interventional masking on Dreaddit, where Channel C is
statistically unaffected by full deletion of content
words ($\Delta = -0.006$, 95\% CI $[-0.037, +0.023]$,
$p=0.72$), suggest the effect is not reducible to
ordinary platform shift; a controlled within-platform
dual-annotation study remains necessary for causal
identification. \tss{} is therefore an audit workflow
that flags label-source-specific shortcut learning
before generalization claims are made.

\section*{Limitations}
\label{sec:limitations}

\paragraph{Data and annotation.}
Evaluation is restricted to four English datasets and to a
binary stress label. For Twitter-gold, inter-annotator
agreement (Cohen's $\kappa$) was not computed under the
manual-annotation protocol available to us; this limits
claims about ``human annotation'' as a general construct.
Twitter-auto inherits a disorder$\rightarrow$stress
conflation from \citet{coppersmith2015adhd}, which is itself
a form of distant-supervision noise.

\paragraph{Confounding between label source and platform.}
The four datasets cross two label sources with two
platforms but do not exhaust the design. The
platform-stratified Twitter-only \dod{}
(\S\ref{ssec:platform-strat}) removes the most obvious
platform confound, but a fully controlled within-platform
dual-annotation experiment is required for causal
identification of label-source effects.

\paragraph{Feature-level ablation and masking control.}
Per-cell bootstrap CIs for the masking suite are now
reported (Tables~\ref{tab:masking}--\ref{tab:masking-deltas}).
We do not perform a feature-level within-channel ablation:
\tss{} is presented as a family-level diagnostic
decomposition, and a full feature-attribution study is left
for future work. We treat feature-family ablation as a
separate attribution study rather than a camera-ready
addition, because the central claim concerns channel-level
label-source divergence rather than ranking individual
features within Channel C. This does not affect the
channel-level claim tested here, but it limits feature-level
interpretability inside Channel C. The \texttt{random\_pos}
negative control does not collapse to chance because POS
distributional cues remain after shuffling.

\paragraph{Word- vs.\ character-level lexical baseline.}
We did not include a word-unigram/bigram lexical baseline
in this camera-ready version; Channel A is intentionally
character-based to stress-test lexical surface dependence
under noisy social-media spelling. Future work should
compare word- and character-level lexical baselines
directly to verify that the \emph{lexical interference
effect} is not specific to sub-word tokenization.

\paragraph{Clinical translation.}
Social-media data are not clinical data. We make no claim
that \tss{} is a diagnostic medical tool; any deployment in
a clinical setting requires clinician oversight, informed
consent, and risk management. \dod{} may also require
calibration for morphologically richer languages
(e.g., Arabic, Chinese), where surface-form variation
interacts differently with character-level lexical
representations.

\paragraph{Few-shot baseline.}
LLaMA-3-8B is reported only in a 3-shot diagnostic
capacity; instability is interpreted as evidence about
few-shot lexical priors, not as a comment on fine-tuned
LLM performance.

\section*{Ethical Considerations}

This work is not a diagnostic medical tool. It focuses on
auditing label bias and improving methodological rigor in
CMH NLP. All evaluations use previously collected research
datasets or released artifact-level data; no new user data
was collected for this study. We use qualitative examples
only after redacting personally identifying details. Any
deployment of distress-detection systems built on
social-media text requires clinician oversight, informed
consent, and careful risk management, particularly given
documented disparities in CMH model behavior across
demographic groups \citep{aguirre2021gender}.

\bibliography{references}

\appendix

\section*{Appendices}
\noindent Sections~A--F below provide mathematical formulations,
full statistical tables, robustness audits, qualitative case
analysis, extended metrics, and SHAP audits referenced in the
main text.

\section{Mathematical Formulations}
\label{app:formulas}

\subsection{Degree of Divergence}

\begin{equation}
\begin{aligned}
\dod \;=\; & \big[M_h(BC) - M_h(A)\big] \\
&\;-\; \big[M_a(BC) - M_a(A)\big]
\end{aligned}
\label{eq:dod}
\end{equation}

\dod{} is a difference-in-differences statistic adapted
from causal econometrics; $M_h$ and $M_a$ denote
\macroF{} under human and auto label sources respectively.
Inference is by instance-level bootstrap
($n{=}10{,}000$).

\subsection{Adaptive Length Smoothing (Channel B)}

\begin{equation}
\lambda \;=\; \lambda_{\min} +
(\lambda_{\max} - \lambda_{\min})\,\min(L/L_{\text{ref}},\,1)
\end{equation}

with $\lambda_{\min}{=}3.0$, $\lambda_{\max}{=}20.0$,
$L_{\text{ref}}{=}50.0$. Short texts retain sparse POS
signals; long texts are regularized.

\subsection{Length-Normalized Log-Odds Mass (Channel B)}

\begin{equation}
B_{\text{pos\_mass}} \;=\;
\frac{\log(1 + M_{\text{pos}})}{\log(1 + L_{\text{eff}})},
\qquad L_{\text{eff}} = L + \lambda
\end{equation}

\noindent The remaining Channel B features are
$B_{\text{polarity}} = M_{\text{pos}} / (M_{\text{pos}} +
M_{\text{neg}} + \varepsilon)$,
$B_{\text{load}} = \log_e(1 + M_{\text{pos}} +
M_{\text{neg}}) / \max(\log_e(1 + \text{cnt}),\, 1)$, and
$B_{\text{abs}} = |\{t : t \in \mathcal{A}\}| /
L_{\text{eff}}$, where $\mathcal{A}$ is a closed
absolutist-word list \citep{almosaiwi2018absolute}. Of the
six Channel B features, $B_{\text{abs}}$ is the only one
that accesses lexical tokens.

\subsection{Structural Negation Balance (Channel C)}

\begin{equation}
\text{NegBalance} \;=\;
\frac{R_{\text{Neg}\rightarrow\text{Verb}} -
      R_{\text{Neg}\rightarrow\text{Adj}}}
     {N_{\text{neg\_verb}} + N_{\text{neg\_adj}} +
      \varepsilon}
\end{equation}

\noindent Operationalizes the helplessness vs.\ negative
self-evaluation distinction motivated by
\citet{beck1976cognitive}.

\subsection{Length-Robust Squashing (Channel C)}

\begin{equation}
c_{\text{out}} \;=\; \tanh(2c)
\end{equation}

\subsection{Length-Independent Lexical Diversity}

\begin{equation}
I \;=\; \frac{M_1^{\,2}}{M_2 - M_1}
\end{equation}

\noindent where $M_1$ is the vocabulary size and
$M_2 = \sum_i f_i^{\,2}$ is the sum of squared per-type
frequencies. This formulation follows the convention used
in our pipeline (\texttt{features.py},
\texttt{\_calc\_yule\_i}); values are capped at
$10^{3}$ to prevent extreme outliers from very short
texts. Yule's $I$ is preferred over TTR because TTR decays
mathematically with text length, biasing cross-platform
comparison from long Reddit posts to short tweets.

\section{Detailed Statistics}
\label{app:stats}

\subsection{SSI Saturation Test (Reddit-combi)}
\label{app:saturation}

\begin{table}[h]
\centering
\footnotesize
\setlength{\tabcolsep}{4pt}
\begin{tabular}{lcc}
\toprule
\textbf{Setting} & \textbf{$\Delta$F1 (BC$-$C)} & \textbf{$p$} \\
\midrule
Reddit-combi (Auto) & $+0.0080$ & 0.259 \\
\bottomrule
\end{tabular}
\caption{SSI signal-saturation equivalence test
(paired instance bootstrap).}
\label{tab:saturation}
\end{table}

\subsection{Representation Efficiency Note}

The single-row paired-bootstrap result for TSS-C vs.\
MentalBERT on Twitter-gold ($\Delta\macroF{}=-0.0141$,
95\% CI $[-0.037, +0.008]$, $p=0.108$) appears in the
top row of Table~\ref{tab:baselines-paired}. The CI does
not exclude zero, but LLaMA-3 (3-shot) reaches \macroF{}
0.800 on the same dataset, so this is \emph{not} a SOTA
claim.

\subsection{Paired Baseline Comparisons}
\label{app:baselines}

\begin{table}[h]
\centering
\footnotesize
\setlength{\tabcolsep}{3pt}
\begin{tabular}{lcc}
\toprule
\textbf{Pair (dataset)} & \textbf{$\Delta$F1} & \textbf{95\% CI} \\
\midrule
C\,/\,MB (Tw-G)       & $-0.014$ & $[-0.037, +0.008]$ \\
C\,/\,LLaMA (Tw-G)    & $-0.099$ & $[-0.121, -0.078]$ \\
C\,/\,MB (Tw-A)       & $+0.086$ & $[+0.073, +0.099]$ \\
C\,/\,LLaMA (Tw-A)    & $-0.018$ & $[-0.032, -0.004]$ \\
BC\,/\,MB (Tw-G)      & $-0.025$ & $[-0.045, -0.004]$ \\
BC\,/\,LLaMA (Tw-G)   & $-0.110$ & $[-0.132, -0.088]$ \\
BC\,/\,MB (Tw-A)      & $+0.037$ & $[+0.027, +0.048]$ \\
BC\,/\,LLaMA (Tw-A)   & $-0.067$ & $[-0.080, -0.054]$ \\
\bottomrule
\end{tabular}
\caption{Paired baseline comparisons (paired bootstrap CI).
Pairs are TSS channel vs.\ baseline. MB~=~MentalBERT;
LLaMA~=~LLaMA-3-8B (3-shot); Tw-G~=~Twitter-gold;
Tw-A~=~Twitter-auto.}
\label{tab:baselines-paired}
\end{table}

\begin{table}[h]
\centering
\footnotesize
\setlength{\tabcolsep}{3pt}
\begin{tabular}{lcc}
\toprule
\textbf{Pair} & \textbf{$p_{\text{McN}}$} & \textbf{$d_z$} \\
\midrule
TSS-C  vs.\ MB (Tw-gold)        & $4.49\!\times\!10^{-5}$    & $-1.47$ \\
TSS-C  vs.\ LLaMA-3 (Tw-gold)   & $2.55\!\times\!10^{-15}$   & $-11.40$ \\
TSS-C  vs.\ MB (Tw-auto)        & $4.49\!\times\!10^{-13}$   & $+14.77$ \\
TSS-C  vs.\ LLaMA-3 (Tw-auto)   & $5.15\!\times\!10^{-4}$    & $-2.83$ \\
TSS-BC vs.\ MB (Tw-gold)        & $1.81\!\times\!10^{-4}$    & $-2.54$ \\
TSS-BC vs.\ LLaMA-3 (Tw-gold)   & $5.11\!\times\!10^{-13}$   & $-12.47$ \\
TSS-BC vs.\ MB (Tw-auto)        & $4.32\!\times\!10^{-3}$    & $+6.56$ \\
TSS-BC vs.\ LLaMA-3 (Tw-auto)   & $<10^{-15}$                & $-10.64$ \\
\bottomrule
\end{tabular}
\caption{McNemar $p$ and paired effect size $d_z$.
MB~=~MentalBERT.}
\label{tab:baselines-mcnemar}
\end{table}

\subsection{Paired Channel Contrasts}
\label{app:contrasts}

\begin{table}[h]
\centering
\footnotesize
\setlength{\tabcolsep}{3pt}
\begin{tabular}{llccc}
\toprule
\textbf{Dataset} & \textbf{Cmp} & \textbf{$\Delta$F1} & \textbf{95\% CI} & \textbf{$p_{\text{perm}}$} \\
\midrule
Dread (H)    & C--A   & $+0.106$ & $[+0.07, +0.15]$ & 0.0001 \\
Dread (H)    & BC--A  & $+0.057$ & $[+0.02, +0.10]$ & 0.0032 \\
Tw-gold (H)  & C--A   & $+0.039$ & $[+0.02, +0.06]$ & 0.0005 \\
Tw-gold (H)  & BC--A  & $+0.029$ & $[+0.00, +0.05]$ & 0.0131 \\
Tw-auto (A)  & C--A   & $-0.019$ & $[-0.03, -0.00]$ & 0.9877 \\
Tw-auto (A)  & BC--A  & $-0.067$ & $[-0.08, -0.05]$ & 1.0000 \\
Red-c. (A)   & C--A   & $+0.070$ & $[+0.04, +0.10]$ & 0.0001 \\
Red-c. (A)   & BC--A  & $+0.078$ & $[+0.05, +0.11]$ & 0.0001 \\
\bottomrule
\end{tabular}
\caption{Paired channel contrasts ($\Delta\macroF{}$) by
dataset, with permutation $p$-values.}
\label{tab:contrasts}
\end{table}

\section{Robustness Audits}
\label{app:robustness}

\subsection{LODO}
\label{app:lodo}

\begin{table}[h]
\centering
\small
\begin{tabular}{lcccc}
\toprule
\textbf{Channel/Model} & \textbf{Mean} & \textbf{Std} & \textbf{Min} & \textbf{Max} \\
\midrule
TSS-A      & 0.662 & 0.042 & 0.628 & 0.723 \\
TSS-B      & 0.570 & 0.045 & 0.531 & 0.637 \\
TSS-BC     & 0.679 & 0.032 & 0.653 & 0.734 \\
TSS-C      & 0.685 & 0.015 & 0.666 & 0.700 \\
MentalBERT & 0.818 & 0.030 & 0.772 & 0.851 \\
\bottomrule
\end{tabular}
\caption{Leave-one-domain-out cross-stressor robustness
on Dreaddit (out-of-domain \macroF{}, mean over five
domains).}
\label{tab:lodo}
\end{table}

\begin{table}[h]
\centering
\small
\begin{tabular}{lcccc}
\toprule
\textbf{Stressor} & \textbf{TSS-A} & \textbf{TSS-C} & \textbf{TSS-BC} & \textbf{MB} \\
\midrule
abuse      & 0.639 & 0.672 & 0.653 & 0.772 \\
anxiety    & 0.723 & 0.700 & 0.734 & 0.828 \\
financial  & 0.632 & 0.690 & 0.683 & 0.833 \\
ptsd       & 0.686 & 0.695 & 0.669 & 0.851 \\
social     & 0.628 & 0.666 & 0.659 & 0.806 \\
\midrule
Mean       & 0.662 & 0.685 & 0.679 & 0.818 \\
Std        & 0.042 & 0.015 & 0.032 & 0.030 \\
\bottomrule
\end{tabular}
\caption{Per-stressor LODO breakdown
(MentalBERT~=~MB).}
\label{tab:lodo-detail}
\end{table}

\subsection{Phenotyping}

\begin{table}[h]
\centering
\footnotesize
\setlength{\tabcolsep}{4pt}
\begin{tabular}{lcc}
\toprule
\textbf{Statistic} & \textbf{Value} & \textbf{95\% CI / $p$} \\
\midrule
Bootstrap ARI         & 0.980 & $[0.961, 0.991]$ ($n_{\text{boot}}{=}100$) \\
Cluster--dataset AMI  & 0.281 & $p{=}0.0005$ (perm 2000) \\
Cluster--label AMI    & 0.084 & $p{=}0.0005$ (perm 2000) \\
\bottomrule
\end{tabular}
\caption{K-Means ($K{=}3$) on Channel C ($N{=}12{,}924$,
$d{=}154$). DoD analysis uses $N{=}12{,}906$;
the 18-instance gap reflects zero-length POS sequences
excluded from \dod{} but retained in clustering via
Channel C features.}
\label{tab:phenotype}
\end{table}

\subsection{Channel Orthogonality}

\begin{table}[h]
\centering
\small
\begin{tabular}{lc}
\toprule
\textbf{Summary} & \textbf{Value} \\
\midrule
Mean $|\rho|$ across pairs    & 0.595 \\
Orthogonality ratio           & 0.095 \\
A vs.\ B (key pair)           & 0.296 \\
Effective $N$                 & 271{,}026 \\
\bottomrule
\end{tabular}
\caption{Instance-level Spearman $\rho$ across channel
pairs.}
\label{tab:orthogonality}
\end{table}

\subsection{Efficiency}
\label{app:efficiency}

\begin{table}[h]
\centering
\small
\begin{tabular}{lcccc}
\toprule
\textbf{Channel} & \textbf{$d$} & \textbf{Train (s)} & \textbf{Reg.} & \textbf{$\ell_1$-ratio} \\
\midrule
A   & 500 & 17.6   & $L_2$       & --- \\
B   & 6   & 1007.9 & ElasticNet  & 0.95 \\
C   & 154 & 27.5   & $L_2$       & --- \\
BC  & 160 & 335.9  & ElasticNet  & 0.95 \\
\bottomrule
\end{tabular}
\caption{Efficiency profile. Channel B's training time is
an artifact of an exhaustive ElasticNetCV grid; once fitted,
inference is near-instantaneous.}
\label{tab:efficiency}
\end{table}

\subsection{LLaMA-3 Few-Shot Instability}
\label{app:llama}

\begin{table}[h]
\centering
\footnotesize
\setlength{\tabcolsep}{4pt}
\begin{tabular}{lcccc}
\toprule
\textbf{Dataset} & \textbf{F1} & \textbf{Prec.} & \textbf{Rec.} & \textbf{Pattern} \\
\midrule
Dreaddit-test (H) & 0.613 & 0.606 & 0.984 & Over-firing \\
Twitter-auto (A)  & 0.522 & 0.846 & 0.203 & Under-firing \\
Twitter-gold (H)  & 0.800 & 0.780 & 0.680 & Balanced \\
\bottomrule
\end{tabular}
\caption{LLaMA-3-8B (3-shot) precision--recall instability
across label sources.}
\label{tab:llama-instability}
\end{table}

\section{Qualitative Conflict Zone}
\label{app:qualitative}

We extracted 800 instances (200 per dataset) for
instance-level qualitative analysis of cases where the
lexical baseline (Channel A) and the structural probes
(Channels B/C) disagree. The full workbook
(\texttt{error\_analysis\_qualitative.xlsx}) is released
with the code.

\paragraph{Workbook statistics.}
The \texttt{Shift\_Cases\_AvsB} sheet documents 682
instances where Channel B corrects Channel A's errors:
415 (60.9\%) are A-false-positives (keyword-driven false
positives) and 267 (39.1\%) are A-false-negatives
(implicit stress). The complementary
\texttt{Reverse\_Shift} sheet documents 667 instances
where A corrects B; restricting to human-labeled data
yields 319 A-corrections (Twitter-gold: 200;
Dreaddit-test: 119), comprising 177 B-false-positives and
142 B-false-negatives. The \texttt{All\_Channels\_Compare}
sheet shows that Channel C alone corrects 253 of A's
errors (195 keyword-driven false positives + 58 implicit
stress), and BC corrects 241.

Table~\ref{tab:qualitative} gives 13 representative cases
illustrating the two failure modes; the full
characterization is in the released workbook.

\begin{table*}[t]
\centering
\small
\renewcommand{\arraystretch}{1.15}
\begin{tabular}{p{2.4cm} p{6.8cm} c c c}
\toprule
\textbf{Case type} & \textbf{Sample (truncated)} & \textbf{GT} & \textbf{A} & \textbf{TSS} \\
\midrule
Keyword-driven FP (advert)         & ``call us at +971 4\,324\,3244 to consult \dots\ \#MentalHealth \#Stress \#DryEye'' & 0 & 1 & 0 \\
Implicit stress (suicidality)      & ``I am on the edge of committing\dots\ a few hours remaining\dots'' & 1 & 0 & 1 \\
Boundary collapse (HR narrative)   & ``I man the front desk\dots\ HR Customer Service Representative\dots'' & 0 & 1 & 0 \\
Keyword-driven FP (resilience)     & ``Resilience is not a trait\dots\ \#mentalhealth \#mentalstrength'' & 0 & 1 & 0 \\
Keyword-driven FP (counseling ad)  & ``Sometimes love is complicated. Let's talk it out. \#counseling \#mentalhealth'' & 0 & 1 & 0 \\
Keyword-driven FP (product ad)     & ``New Product Alert! Silicone Suction Snapper. Use code SOCIAL10 \#Stress'' & 0 & 1 & 0 \\
Keyword-driven FP (achievement)    & ``Squatting 315\,lbs for 6 reps\dots\ Tonight I just did 305!!!! I'm so excited!!!'' & 0 & 1 & 0 \\
Keyword-driven FP (policy)         & ``When moving into their tiny house, they would be given a state I.D.\dots'' & 0 & 1 & 0 \\
Implicit stress (medical trauma)   & ``\dots they needed to strap me down\dots\ I completely dissociated.'' & 1 & 0 & 1 \\
Implicit stress (relational)       & ``The one person you thought you could talk to blows you off. tired alone hurt'' & 1 & 0 & 1 \\
Implicit stress (masked depression)& ``feel sad all the time then happy for a moment\dots\ faking it for days'' & 1 & 0 & 1 \\
Implicit stress (burnout)          & ``stress at work\dots\ 60 hours a week (even when I had Covid)\dots'' & 1 & 0 & 1 \\
Implicit stress (financial crisis) & ``\dots paid 1{,}000 dollars for hospital bills\dots\ now we have nothing left.'' & 1 & 0 & 1 \\
\bottomrule
\end{tabular}
\caption{Representative qualitative conflict-zone cases.
GT~=~ground truth, A~=~Channel A, TSS~=~Channel B/C
combined verdict. The first eight cases are
keyword-driven false positives where A fires on topical
markers; the last five are implicit-stress cases where A
misses distress expressed without explicit stress
vocabulary.}
\label{tab:qualitative}
\end{table*}

\section{Extended Metric Dashboard}
\label{app:metrics}

\begin{table}[h]
\centering
\footnotesize
\setlength{\tabcolsep}{2pt}
\begin{tabular}{llccccc}
\toprule
\textbf{Data} & \textbf{Ch.} & \textbf{F1} & \textbf{PR-AUC} & \textbf{MCC} & \textbf{B.Acc} & \textbf{Prev} \\
\midrule
Dread (H) & A  & 0.634 & 0.809 & 0.324 & 0.646 & 0.516 \\
Dread (H) & C  & 0.739 & 0.823 & 0.495 & 0.740 & 0.516 \\
Dread (H) & BC & 0.690 & 0.819 & 0.421 & 0.697 & 0.516 \\
Tw-G (H)  & A  & 0.661 & 0.540 & 0.342 & 0.680 & 0.338 \\
Tw-G (H)  & C  & 0.701 & 0.637 & 0.422 & 0.689 & 0.338 \\
Tw-G (H)  & BC & 0.690 & 0.708 & 0.438 & 0.677 & 0.338 \\
Tw-A (A)  & A  & 0.522 & 0.510 & 0.065 & 0.531 & 0.478 \\
Tw-A (A)  & C  & 0.504 & 0.656 & 0.209 & 0.568 & 0.478 \\
Tw-A (A)  & BC & 0.455 & 0.658 & 0.167 & 0.545 & 0.478 \\
Red-c (A) & A  & 0.611 & 0.951 & 0.223 & 0.608 & 0.880 \\
Red-c (A) & C  & 0.681 & 0.964 & 0.371 & 0.710 & 0.880 \\
Red-c (A) & BC & 0.689 & 0.968 & 0.380 & 0.702 & 0.880 \\
\bottomrule
\end{tabular}
\caption{Extended metric dashboard. Tw-G~=~Twitter-gold,
Tw-A~=~Twitter-auto. \macroF{} is the headline metric;
PR-AUC is the imbalance-sensitive complement; MCC and
balanced accuracy are further robustness checks under
skewed prevalence.}
\label{tab:dashboard}
\end{table}

\section{SHAP Lexical Concentration}
\label{app:shap}

\begin{table}[h]
\centering
\footnotesize
\setlength{\tabcolsep}{4pt}
\begin{tabular}{lccc}
\toprule
\textbf{Mode} & \textbf{$n$} & \textbf{Mean conc.} & \textbf{95\% CI} \\
\midrule
lexical\_wins & 40 & 0.333 & $[0.308, 0.355]$ \\
\bottomrule
\end{tabular}
\caption{Top-10 SHAP concentration on Channel A
\citep{lundberg2017unified} when lexical predictions beat
structural ones. The top 10 features (2\% of the
500-feature space) capture 33.3\% of total absolute SHAP
weight ($16\!\times$ the uniform-contribution
expectation), consistent with strong shortcut reliance in
SSI-rewarded regimes.}
\label{tab:shap}
\end{table}

\end{document}